\documentclass[preprint,12pt]{elsarticle}

\usepackage{amssymb}
\usepackage{caption}
\usepackage{subcaption}
\usepackage[ruled,vlined]{algorithm2e}
\usepackage{float}
\usepackage{bbm}

\journal{Composites Science and Technology}

\begin{document}

\begin{frontmatter}



\title{Automated processing of X-ray computed tomography images via panoptic segmentation for modeling woven composite textiles}


\author{Aaron Allred$^{a,b}$}
\author{Lauren J. Abbott $^b$}
\author{Alireza Doostan\corref{cor1} $^a$}
\cortext[cor1]{Corresponding author; Email, alireza.doostan@colorado.edu; Phone, (303) 492-7572}
\author{Kurt Maute $^a$}

\affiliation{organization={Smead Aerospace Engineering Sciences, University of Colorado},
            addressline={3775 Discovery Dr.}, 
            city={Boulder},
            postcode={80303}, 
            state={Colorado},
            country={United States}}
            
\affiliation{organization={Thermal Protection Materials Branch, NASA Ames Research Center}, 
            city={Moffett Field},
            postcode={94035}, 
            state={California},
            country={United States}}

\begin{abstract}
A new, machine learning-based approach for automatically generating 3D digital geometries of woven composite textiles is proposed to overcome the limitations of existing analytical descriptions and segmentation methods.  In this approach, panoptic segmentation is leveraged to produce instance segmented semantic masks from X-ray computed tomography (CT) images. This effort represents the first deep learning based automated process for segmenting unique yarn instances in a woven composite textile. Furthermore, it improves on existing methods by providing instance-level segmentation on low contrast CT datasets. Frame-to-frame instance tracking is accomplished via an intersection-over-union (IoU) approach adopted from video panoptic segmentation for assembling a 3D geometric model.  A corrective recognition algorithm is developed to improve the recognition quality (RQ).  The panoptic quality (PQ) metric is adopted to provide a new universal evaluation metric for reconstructed woven composite textiles. It is found that the panoptic segmentation network generalizes well to new CT images that are similar to the training set but does not extrapolate well to CT images of differing geometry, texture, and contrast. The utility of this approach is demonstrated by capturing yarn flow directions, contact regions between individual yarns, and the spatially varying cross-sectional areas of the yarns.

\end{abstract}




\end{frontmatter}


\section{Introduction}
\label{Intro}

\subsection{Background}

Woven composite textiles have emerged as beneficial building blocks for developing high performance, low weight structures in the field of aerospace. However, structures composed of composite material systems have traditionally relied on empirical analysis techniques to determine mechanical and thermal performance \cite{hill,hashin,christensen}. To reduce the scope of required empirical testing, there is a need to expand upon the current computational capabilities when modeling composite material systems. Still developing, multiscale analysis of composite material systems seeks to determine various macroscopic material properties by accounting for heterogeneities and spatially varying uncertainties across multiple length scales \cite{feyel,kanoute,bostanabad}. Herein, the mesoscale refers to the intermediate scale of a woven composite textile where the individual weave yarns are distinguishable. At this scale, the microscopic constituents (e.g. individual fibers) are considered to be homogeneous.  

The resultant material properties and mechanical and thermal performance of woven composite textiles on the macroscale are heavily influenced by the weave geometry at the mesoscale. For this reason, finite element analysis (FEA) of mesoscale geometries is commonly conducted. However, the quality of the performance characteristic being quantified is dependent on the fidelity of the model's geometry. This is particularly true for woven composite textiles, which have been studied in great depth \cite{adanur,kuhn,ansar,dasgupta}.  For modeling these material systems, a common practice is to construct composite weave mesostructures via analytical functions which characterize the geometry \cite{lomov1, lomov2, lomov3, badel, wendling, ha}.  These analytical descriptions are based on observations of the yarn orientation and result in idealized geometries which fail to account for variations in the geometry (e.g. individual yarn cross-sectional areas and yarn path detours). One such popular open-source tool for modeling an idealized composite weave geometry at the mesoscale is TexGen \cite{TexGen}.

Efforts have emerged to improve the analytical descriptions of composite weave geometries at the mesoscale.  A consistent geometry generating procedure which considers warping yarn cross-sections due to contact was developed to increase the fidelity of the geometry for FEA \cite{wendling}.  Moreover, there is a desire to move away from deterministic predictions of the macroscopic material properties towards stochastic modeling for uncertainty quantification.  Recently, randomness has been imposed on empirically defined distributions to increase the fidelity of mechanical models of composite weaves \cite{ji}. In all of these approaches the modeling quality of the mesoscale geometry is limited by the variations identified by the operator and thus features a human-in-the-loop component. 

In order to address these shortcomings, X-ray computed tomography aided engineering (XAE) approaches have emerged.  XAE seeks to model material systems directly from X-ray computed tomography (CT) data \cite{auenhammer, emerson,kalender}. Constructing 3D models for mechanical analysis from CT imaging provides the capability of modeling variations in the material system seen during production instead of relying on idealized geometries to determine performance. The bottleneck for these approaches lies in segmenting the images. Image segmentation (or segmentation in general) refers to the process of partitioning an image into multiple segments often by assigning a pixel identifier. At the mesoscale, an image of multiple yarns may be segmented by assigning unique pixel identifiers to each yarn. Manually segmenting a dataset is undesirable due to the labor requirement and inherent inaccuracies from human-error.

Various image processing techniques have been developed for automatically segmenting woven composites to avoid manual segmentation. The first major development came when Naouar et al. applied a structure tensor methodology to achieve segmentation \cite{naouar1}. Because this methodology struggles with complex 3D textiles such as those with a binding thread, textile analysis of the Grey Level Co-occurrence matrix (GLC Matrix) has been employed as an alternative \cite{naouar2}. In a recent study by Huang et al. \cite{huang}, where a manual segmentation procedure is proposed, it is noted that none of these methods provide a way to check the CT-based geometry for consistency with the original geometry after reconstruction.  The most recent procedure for automatically segmenting a composite material system for FEA relies on utilizing the commercial software Azivo for thresholding the CT data \cite{auenhammer}. Thresholding segmentation relies on separating objects within an image based on their greyscale values.  This method is incapable of automatically distinguishing between warp and weft yarns as each yarn in a class (warp and weft) have similar greyscale values. In image processing, a class is a group of like objects in an image. While the authors report accurate segmentation on non-crimped fabric reinforced composites, thresholding is not necessarily transferable to CT scans of other material systems as manual parameters must be set depending on the composite architecture.  Furthermore, the authors report their procedure to be computationally expensive during segmentation.  All these methods rely on the intensity to which the yarns differ and thus are not guaranteed to be generalizable to CT scans with low contrast or textiles with complex geometries. 

To circumvent these limitations, a trained deep learner can be leveraged to automatically segment a composite weave material system.  Badran et al. implement a semantic segmentation methodology utilized for the purpose of defect detection in fiber reinforced ceramic composites. Semantic segmentation was achieved via the commercial software Dragonfly \cite{badran}.  Semantic segmentation identifies classes within an image by assigning a unique pixel identifier to the class (eg. warp yarns assigned blue pixels and weft yarns assigned red pixels) \cite{long,ronneberger}. It is useful for detecting abnormalities such as flaws. Similarly, Kopp et al. leverage semantic segmentation to identify matrix damage in composites as a method of non-destructive evaluation \cite{kopp}.  For the purpose of modeling woven composites, Sinchuk utilized semantic segmentation to discern between weft and warp yarns \cite{sinchuk}.  While these approaches are well suited for identifying defects, and in the latter case volume fraction, they are not fit to model composite weave geometries for FEA because semantic segmentation does not identify unique instances of like classes. An instance is each object in a class. For modeling geometries at the mesoscale, each yarn comprises an instance. Therefore, individual yarns cannot be modeled using semantic segmentation. For modeling the mechanical and thermal response of a woven composite textile at the mesoscale, individual instances (each unique yarn) must be identified in order to model contact regions and assign anisotropic constitutive properties to the yarns. In the field of computer vision, the identification of unique instances via deep learning can be accomplished via instance segmentation tools \cite{he,bai,liu}. Instance segmentation identifies each unique object within an image by assigning labels to each object often with a bounding box around each object.

Combining the architectures of an instance segmentation and semantic segmentation network, panoptic segmentation is a deep learning network capable of identifying each unique instance in a class while also supplying each instance with a unique pixel identifier (i.e. mask) \cite{kirillov}. When conducting multiscale FEA or simply FEA at the mesoscale, the microscale details (e.g. individual fibers) are often not desired in the mesoscale model.  Instead, the individual fibers are often realized on a separate microscopic model \cite{feyel, bostanabad}. Thus, uniform masks of each yarn without each individual fiber details are desirable, and this is made achievable via panoptic segmentation. A method of quantifying the accuracy of the instance predictions (yarn masks) from the panoptic segmentation network can be computed via the panoptic quality (PQ) metric \cite{kirillov}. 

\subsection{Objective}
The approach presented in the paper for generating digital representations of composite weave geometries at the mesoscale incorporates a panoptic segmentation neural network to automatically identify yarn classes (weft and warp) as well as individual yarn instances in a CT dataset. As a result, the first deep learning based automated process for segmenting unique yarn instances in a woven composite textile is presented. The PQ metric is adopted to provide a new universal metric for evaluating the consistency of reconstructed woven composite textile geometries at the mesoscale. 

The rest of this manuscript is organized as follows. Section \ref{IAPM} details the panoptic segmentation network as well as the algorithms utilized to both concatenate predictions from 2D frames into 3D yarns and correct the predictions from the neural network. Section \ref{Mat} outlines the material systems utilized to demonstrate this approach. The results are presented in Section \ref{Results} and discussed in Section \ref{Disc}.

\section{Automated Processing Methodology}
\label{IAPM}

The following methodology converts a dataset comprising consecutive images of a material system into a 3D digital geometry.  This is achieved by combining panoptic segmention, which is performed on each 2D image, with an instance tracking algorithm and a corrective recognition algorithm.  The panoptic segmentation network is responsible for identifying and masking individual yarn instances, the instance tracking algorithm is responsible for grouping yarns across images, and the corrective recognition algorithm improves the predictions by examining consecutive images.

\subsection{Panoptic Segmentation}
\begin{figure}[h]
	\centering
	\includegraphics[width=1\textwidth]{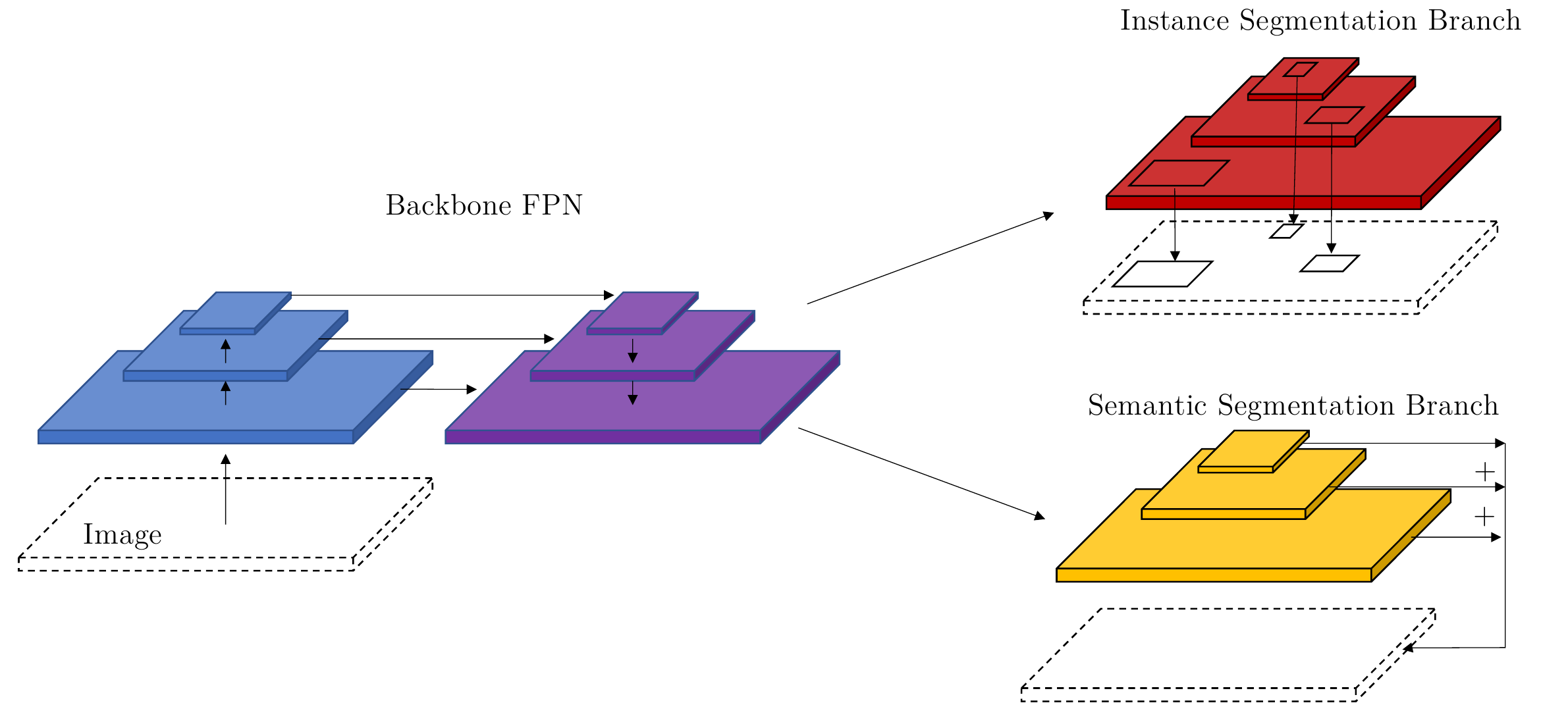}
	\caption{Panoptic segmentation network architecture.}
	\label{fig:pan}
\end{figure}

The panoptic segmentation network outputs a set of instance predictions (e.g. unique masks of each yarn) within each 2D frame, and the network architecture composed by Kirillov et al. is described in Fig. \ref{fig:pan}. The panoptic segmentation network is comprised of two key branches: an instance branch and a semantic branch.  The hyperparameters of the network (e.g. number of convolutional layers, filter sizes, etc.) are dictated by a modular 'backbone' which can be swapped out at will. The instance branch disambiguates like classes by identifying unique instances within a frame; this is tantamount to an instance segmentation network (i.e. a Mask R-CNN network \cite{he}). The semantic branch provides a unique mask to each class in a frame; this is tantamount to a semantic segmentation network.  The learned instances from the instance branch act as stencils to produce unique masks for each instance within a frame. For a woven composite textile, the panoptic segmentation network provides a unique mask for each yarn in an image, thus each yarn cross-section in an image is assigned a unique color.

The loss function utilized during training is a weighted combination of the instance segmentation loss and semantic segmentation loss from each branch. Coined panoptic loss, it is expressed as the following:
\begin{equation}
\mathcal{L}=\lambda_i (\mathcal{L}_{cls}+\mathcal{L}_{bbox}+\mathcal{L}_{mask})+\lambda_s \mathcal{L}_s  .
\end{equation}
The terms $\mathcal{L}_{cls}$ (classification loss), $\mathcal{L}_{bbox}$ (bounding-box regression loss), and $\mathcal{L}_{mask}$ (mask loss) comprise the instance segmentation loss, and $\mathcal{L}_s$ is the semantic segmentation loss. The classification loss quantifies unidentified objects (e.g. object predicted to be the background), the bounding-box regression loss represents the error between the predicted boundary around an object and the boundary around the ground truth, the mask loss is computed as the pixel-to-pixel difference between the ground truth and prediction mask, and the semantic segmentation loss is the difference between the predicted class mask and the ground truth class mask. More information regarding the loss can be found in \cite{kirillov}. The network backbone which dictates the hyperparameters is a ResNet-101-FPN \cite{he2,kirillov2}. The tuning parameters $\lambda_i$ and $\lambda_s$ are predefined by the ResNet-101-FPN network and are not altered in this study. More details on the network backbone are provided in \ref{AppA}. Facebook’s open-source Detectron2 toolset was utilized for training the panoptic segmentation network \cite{wu}. Stochastic gradient descent (SGD) is the optimizer.

\begin{figure}[h]
	\centering
	\includegraphics[width=1\textwidth]{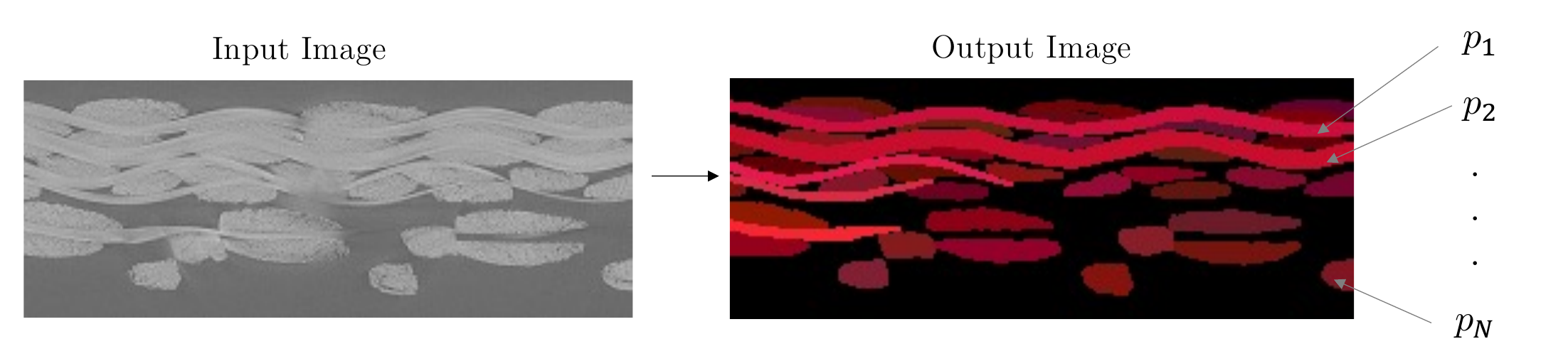}
	\caption{Panoptic segmentation input-output with labeled predictions, $p_1 - p_N$}
	\label{fig:pred}
\end{figure}

An example panoptic segmentation input-output pairing for a woven composite textile is displayed in Fig. \ref{fig:pred}. A non-segmented image is provided, and a 2D frame is generated with uniform masks of each individually identified yarn cross-section. The masks output by the network are 'panoptic predictions,' $p_i$, where $i = 1,...,N$, and $N$ is the number of predictions in a frame. Each panoptic prediction is described by pixels with a unique color identifier.

\subsection{Frame-to-Frame Instance Tracking}

Because the panoptic segmentation network outputs predictions within a 2D frame, a tracking algorithm is required to group yarn cross-sections in consecutive frames into a 3D yarn geometry, $y_i$ where $i = 1,...,Y$, and $Y$ is the number of yarns in a dataset. The goal is to find the yarn, $y_i$, that prediction $p_j$ belongs to such that $p_j \subset y_i$ (thus $y_i$ is a voxel representation). An intersection-over-union (IoU) approach is adopted in order to track predictions.  When tracking objects, the IoU is defined as the ratio of overlapping areas to the total combined area between two instances. This approach is tantamount to video panoptic segmentation \cite{kim,yang}. In this approach, the IoU is evaluated by the relation: 
\begin{equation} \label{IOU}
IoU(p_i,p_j) = \frac{|p_i \cap p_j |}{|p_i \cup p_j|} \; \; \textrm{where} \;\; p_i \in f_k \; \; \textrm{and} \; \; p_j \in f_{k+1} .
\end{equation}
The above relation computes the overlapping areas as the number of shared pixels between two instances ($|p_i \cap p_j |$) in consecutive frames (frame $f_k$ and its consecutive frame $f_{k+1}$). The total combined area between two instances ($|p_i \cup p_j|$) is the total number of unique pixels occupied by the two instances in consecutive frames. The following criteria is considered for evaluating if a prediction $p_j$ should be classified as belonging to the same yarn as prediction $p_i$:
\begin{equation} \label{track}
\textrm{if} \;\; \max_{p_i \in f_k}{IoU(p_i,p_j)} \geq R \;\; \textrm{then} \;\; p_j \in y_i .
\end{equation}
If the IoU meets a threshold requirement, $R$, the prediction, $p_j$, is classified as belonging to the yarn that prediction $p_i$ also belongs ($y_i$). The threshold value is dependent on the material system being scanned and the change in depth of subsequent images. A suitable value of $R$ is provided in the following section as it is dependent on the dataset being evaluated.

\begin{figure}[h]
	\centering
	\includegraphics[width=0.62\textwidth]{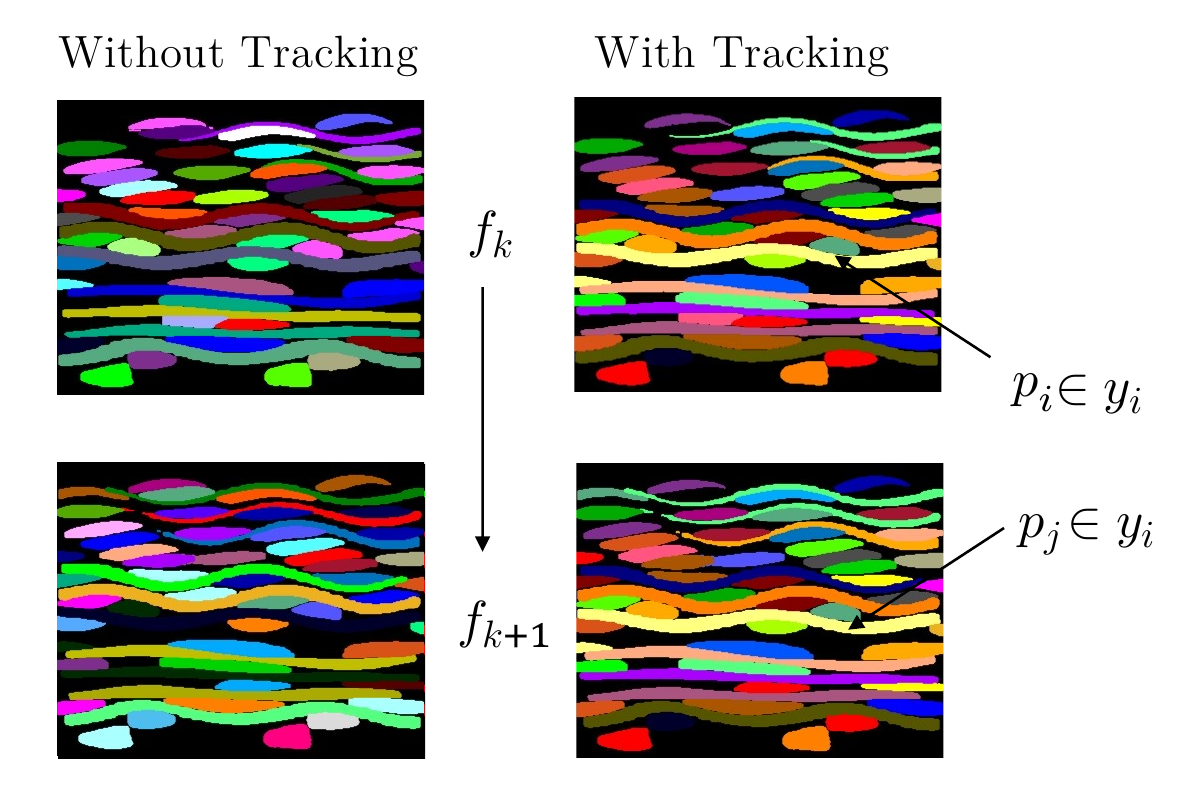}
	\caption{Frame-to-frame instance tracking. The left hand side corresponds to successive panoptic segmentation predictions without tracking.  The right hand side corresponds to panoptic segmentation predictions where like yarn instances are correctly grouped.}
	\label{IoU}
\end{figure}

	To speed up the computation time and accuracy, only previous frame yarn predictions belonging to the same class, weft or warp, are considered.  Using unique color identifiers to classify each yarn instance, Fig. \ref{IoU} demonstrates panoptic segmentation predictions both without and with frame-to-frame instance tracking. The implemented IoU multi-object tracking algorithm is provided below.

\begin{algorithm}[H]
\SetAlgoLined
\textbf{Input:} Set of all frames, $F$, with instance predictions, $p_i$, in each frame, $f_k$, where $k=1,...,|F|$

//Loop over all instances in frame $k+1$\;
 \For{$j=1,...,N_j$}{
   // Loop over all instances over all instances in frame $k$\;
  \For{$i=1,...,N_i$}{
   // Get IoU of instance $j$ and $i$\;
   Compute and store $IoU(p_i,p_j )$\;
   }{
    // Get maximum $IoU$ for instance $j$\;
    Compute $max_{p_i \in f_k} IoU(p_i,p_j)$\;
    // Check if maximum IoU is greater than threshold\;
    \If{$max_{p_i \in f_k} IoU(p_i,p_j) \geq R$}{
    // Assign the instance ID from $p_i$ to $p_j$ such that $p_i,p_j \in y_i$\;
    }
  }
 }
 \caption{Basic procedure for tracking instances across frames}
\end{algorithm}

\subsection{Panoptic Quality}

A clear advantage panoptic segmentation offers over all prior segmentation approaches for composite textiles is the ability to assess the reconstruction of a composite textile via the panoptic quality (PQ) metric in combination with a ground truth image. PQ is defined in \cite{kirillov} as:  

\begin{equation}
PQ = \frac{\sum_{p_i,g_i \in TP} IoU(p_i,g_i)}{|TP|+ \frac{1}{2} |FP|+|FN|} ,
\end{equation}
where the $IoU$, described by Equation \ref{IOU}, is calculated between a prediction and its ground truth.

The PQ metric compares the predictions against the ground truth considering three outcomes: true positives (TP), false positives (FP), and false negatives (FN). A TP is a correctly identified prediction, a FP is an incorrectly constructed prediction, and a FN is an incorrectly not constructed prediction. For computing the PQ score, the IoU is computed between each prediction, $p_i$, and its respective ground truth instance, $g_i$, for all true positive predictions in a frame. The PQ is constructed as the product of the segmentation quality and the recognition quality. PQ can be decomposed into the following equation:

\begin{equation} \label{RQSQ}
PQ = \frac{\sum_{p.g \in TP} IoU(p,g)}{|TP|} \times \frac{|TP|}{|TP|+ \frac{1}{2} |FP|+|FN|} .
\end{equation}
The segmentation quality (left side of Equation \ref{RQSQ}) is the average IoU of true positive segments in a frame.  The recognition quality (right side of Equation \ref{RQSQ}) penalizes the score for false positives and false negatives present in a given prediction of a frame.  The recognition quality is the same as the F1 score in \cite{van}. The panoptic segmentation architecture implicitly works to optimize the PQ in a deep learning setting by minimizing the loss.  

PQ is useful for determining the consistency of the reconstructed geometry at the mesoscale compared to the ground truth. For assessing the panoptic quality of a 3D model, a simple average over the set of all frames, F, in the prediction dataset is proposed:

\begin{equation}
PQ_{avg} = \frac{1}{|F|} \sum_{f \in F} PQ_f .
\end{equation}

\subsection{Corrective Recognition}
The predictions from the panoptic segmentation network are often not exact (the panoptic quality is typically not 100). Additionally, processing a composite material system is often not restricted by real-time processing constraints.  For these reasons, a post-processing methodology is developed for identifying and removing potential false positives as well as constructing instances to replace false negatives not identified by the panoptic segmentation network.

The proposed post-processing procedure considers the continuous nature of the true yarn geometries.  For the dataset segmented via panoptic segmentation, a regional block of $|B|$ subsequent frames, where $B \subset F$, is assembled. If a prediction $p_i \in y_i$ only occurs once in the regional block and this occurrence is not the first or last frame of the regional block, it is flagged as a false positive. If a prediction occurs more than once in the regional block and is bound by frames where that yarn is also identified, the intermediate frames between occurrences are flagged as false negatives. For each frame in the prediction dataset, a new regional block of size $|B|$ is constructed. A suitable size of $B$ is provided in the following section as it is dependent on the dataset being evaluated.

\begin{figure}[h]
     \centering
     \begin{subfigure}[b]{0.48\textwidth}
     	\centering
     	\includegraphics[scale=0.5]{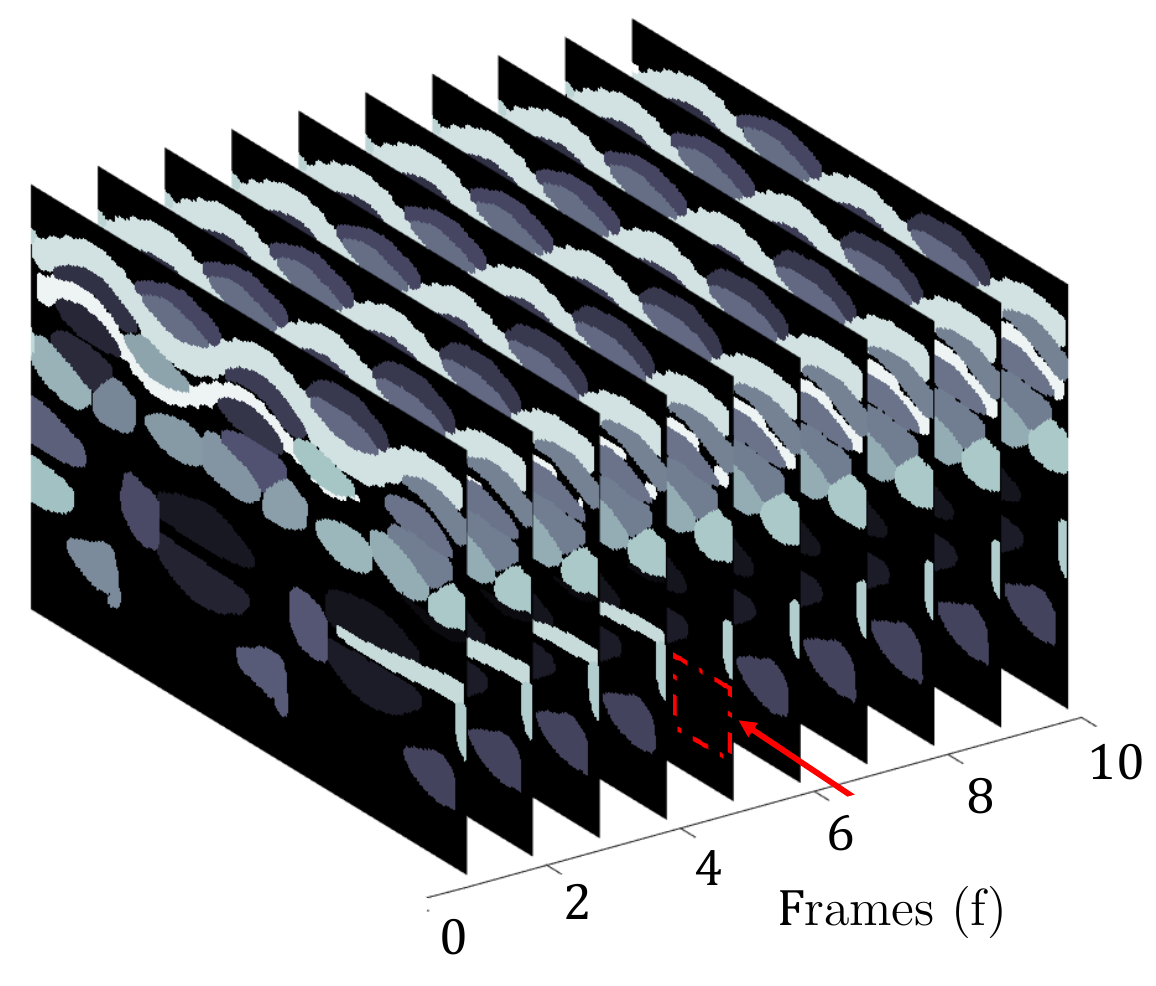}
     	\caption{A FN marked by the dashed red square.}
     \end{subfigure}
     \begin{subfigure}[b]{0.48\textwidth}
     	\centering
     	\includegraphics[scale=0.5]{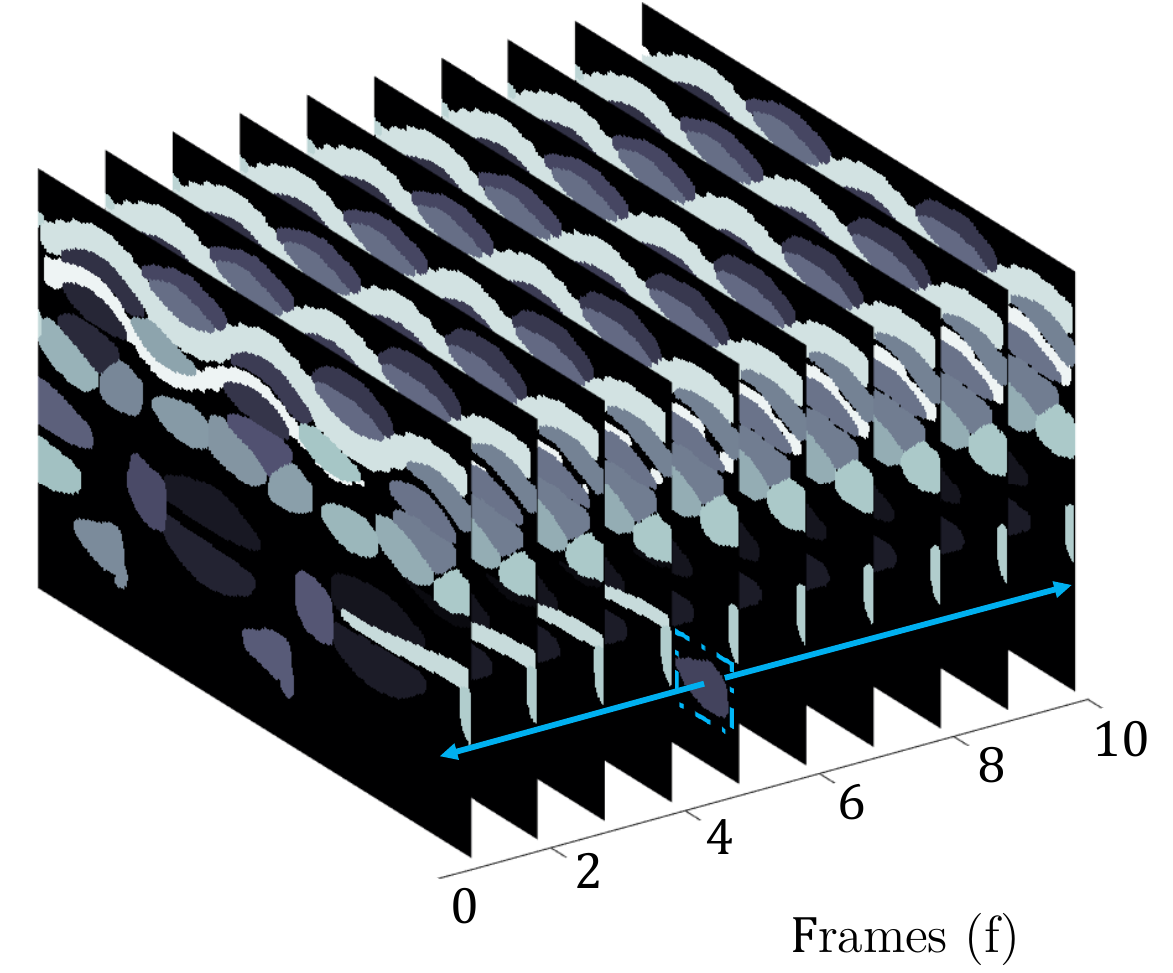}
     	\caption{A FP marked by the dashed blue square.}
     \end{subfigure}
     	\label{Corr}
     	\caption{Regional blocks of size $|B|=10$ are displayed with a false negative present (a) and with a false positive present (b).}
     \label{fig:instanced}
\end{figure}

\begin{figure}[H]
	\centering
	\includegraphics[scale=0.7]{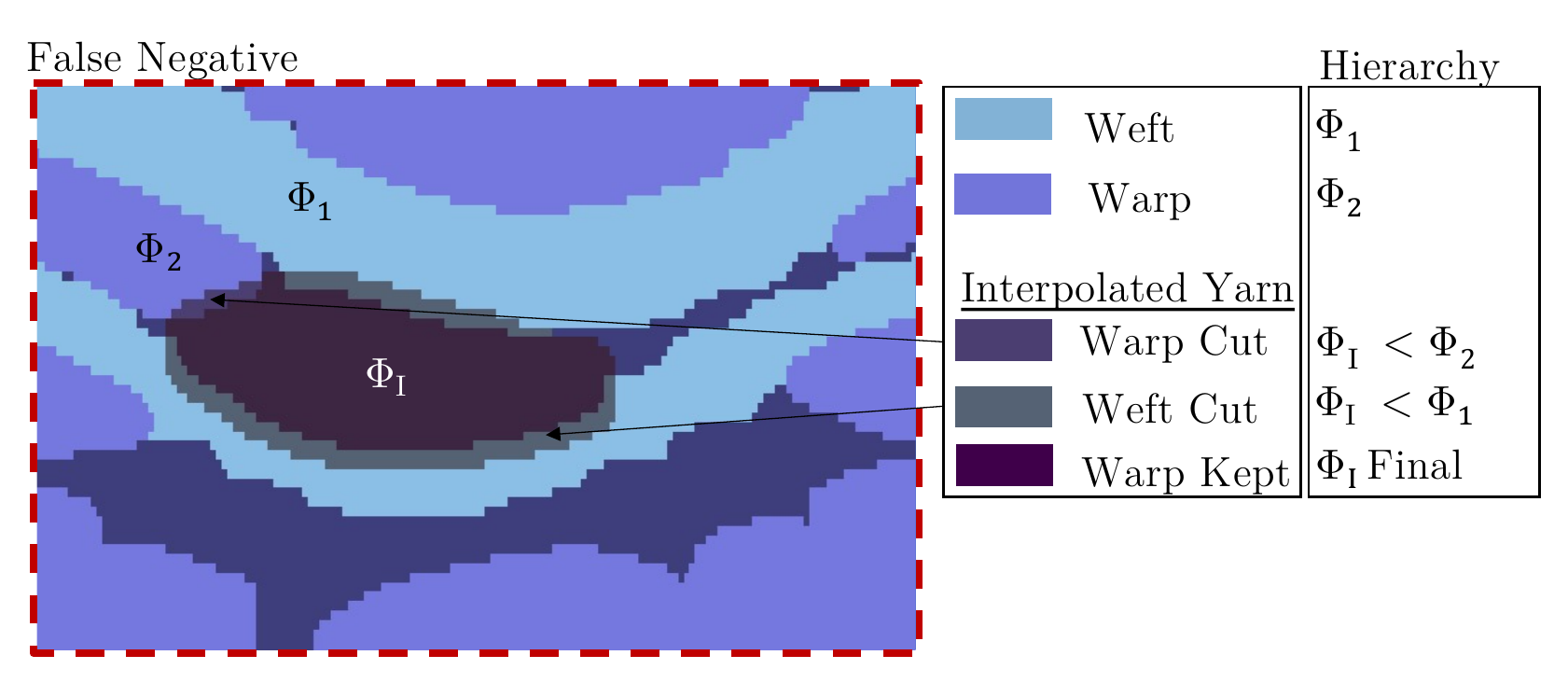}
	\caption{Hierarchical interpolation scheme}
	\label{Interp}
\end{figure}

After identifying the predictions and their frames where false positives occur, the predictions are removed from their frames. Next, after identifying predictions where false negatives occur, interpolation is conducted across frames thus creating an artificial semantic mask to replace the false negative.  A linear hierarchical interpolation scheme is implemented where the existing weft and warp yarn pixels, $\Phi_1$ and $\Phi_2$ respectively, are always granted priority over the interpolation yarn pixels, $\Phi_I$.  A visual of the hierarchical scheme is provided in Fig. \ref{Interp}.

The false positive and false negative identifying algorithm is provided below.

\begin{algorithm}[H]
\SetAlgoLined
\textbf{Input:} Set of frames, $F$, containing tracked yarns, $y_i$
	
// // Loop over all frames, $f_k \in F$ where $k=1,...,|F|$\;
 \For{$k=1,...,|F|$}{
   // Loop over all unique instances, $p_i$, in frame $k$ where $i=1,...,N$\;
 \For{$i=1,...,N$}{
   // Get region block $B$\;
   $B \subset F$ where $B = \{f_{k},...,f_{k+|B|-1}\}$\;
   // Count and store number of occurrences for $y_i$ in block B\;
   $occurrences = \sum_{k=1}^{|B|} \mathbbm{1}_{f_k}(y_i) $\;
   // Store the frame indexes of occurrences
   $f \subset \{k,...,k+|B|-1\}$ \;
   // Check for false positives or false negatives\;
   \If{$ occurrences < |B|$  }{
   // Check if false positive \;   
   \If{$occurences=1 \;\; \textbf{and} \;\; f < (k+|B|-1) \;\; \textbf{and} \;\; f > k$}{
    // Mark instance as false positive\;
    Remove instance prediction $p_i$ from $B$\;
	}
	// Check if false negative\; 
	// Calculate frame index for missing and present occurrences\;
	Missing: $M \subset B$ where $y_i \notin B$\;
	Present: $P \subset B$ where $y_i \in B$\;
	// Check if interpolation should occur for each missing frame, $f_m \in M$ where $m=1,...,|M|$\;
   \For{m = 1,...,$|M|$}{ 
     \If{$m < |M| \;\; \textbf{and}\;\; m > 1$}{
     // Compute and store interpolated geometry $\Phi_I$\;
     	// Assign $\Phi_I$ to where there are no predictions\;
     	\If{ $f_m(\Phi_I) = \emptyset$}{
     	// Assign interpolated geometry\;
     	$f_m(\Phi_I) = \Phi_I$
     	}
     }
     }   
   }
  }
 }
 \caption{Algorithm 2: Corrective procedure to reduce false positives and false negatives}
\end{algorithm}

\section{Automated Processing Example}
\label{Mat}
\subsection{Material Dataset}
The proposed methodology presented in the previous section is general and can be applied to a variety a material systems. In order to demonstrate the feasibility and utility of this proposed approach and to provide a concrete setting for the developed methodology, a dry woven composite textile material system designed for NASA’s Adaptive Deployable Entry and Placement Technology (ADEPT) is examined \cite{semeraro_p1,semeraro_p2}.  Two weave geometries are considered: one with four layers of weft yarns (four ply) and one with six layers of weft yarns (six ply). Both data sets featured a pixel size of 16.1 $\mu m$, and a frame-to-frame difference (depth) of 16.1 $\mu m$. The six ply dataset comprises 1024 consecutive frames with a resulution of 194x500 pixels. The four ply dataset comprises 850 consecutive frames with a resulution of 158x870 pixels. Training, validation, and testing data is sampled uniformly from the six ply weave in a 60\%-20\%-20\% split, respectively. To examine the transferability of the trained network to a similar composite textile with a different mesostructure and different greyscale contrast, the four ply weave was not included in the training.  In addition to the raw CT datasets, manually instance segmented ground truth data is used for training and evaluation (an example of an open source tool for manually instancing a dataset is LabelMe \cite{russell}).

\subsection{Algorithm Parameters}

Considering the instance tracking algorithm, an IoU tracking threshold ($R$ value in Equation \ref{track}) of 0.4 was found to perform well for the ADEPT material system with a 16.1 $\mu m$ change in depth between frames. For the corrective recognition algorithm, a region block size of $|B| = 10$ was found to perform well for this material system.

\begin{figure}
     \centering
     \begin{subfigure}[b]{0.3\textwidth}
         \centering
         \includegraphics[scale = 0.28]{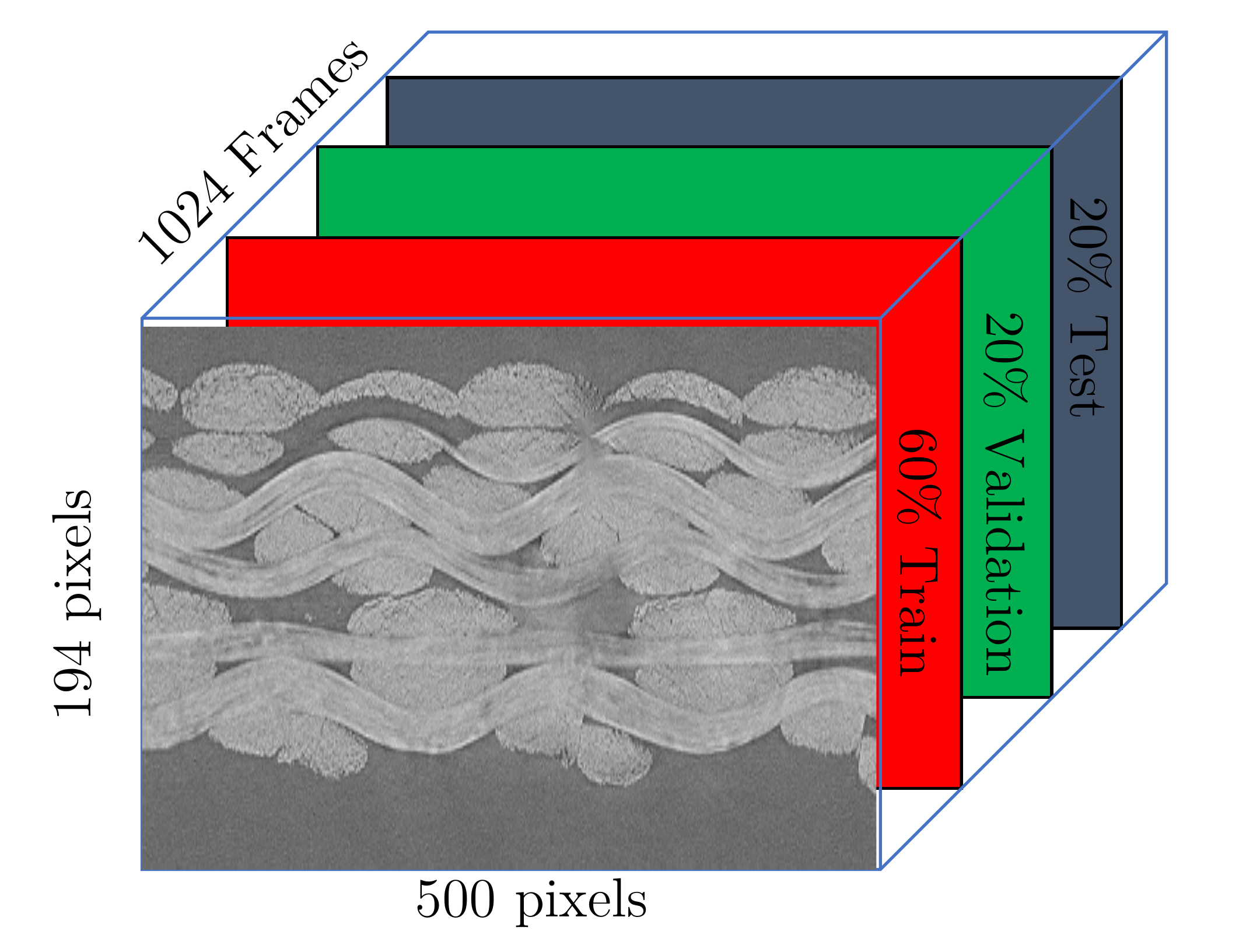}
         \caption{The six ply configuration.}
         \label{Data6}
     \end{subfigure}
     \hspace{20mm}
     \begin{subfigure}[b]{0.5\textwidth}
         \centering
         \includegraphics[scale = 0.35]{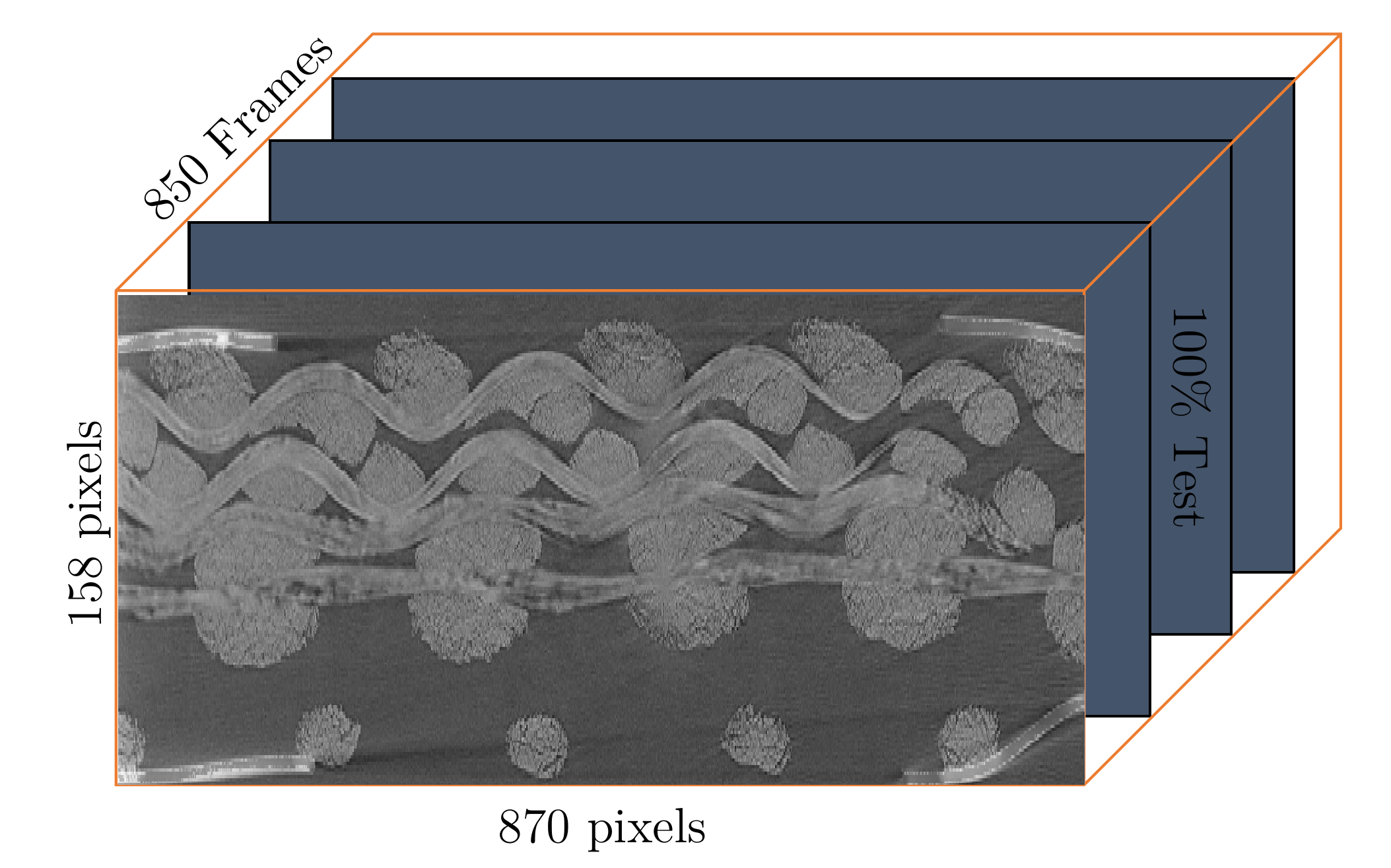}
         \caption{The four ply configuration.}
         \label{Data4}
     \end{subfigure}
        \caption{CT images were received in frame data blocks from NASA Ames Research Center \cite{semeraro_p1,semeraro_p2,vanaer}.  The pixel resolution of each configuration is provided as well as the percent allocations towards training, validation, and testing.}
        \label{fig:Dataset}
\end{figure}

\subsection{Blurred Dataset}

\begin{figure}[h]
     \centering
     \begin{subfigure}[b]{0.45\textwidth}
     	\centering
     	\includegraphics[scale = 0.6]{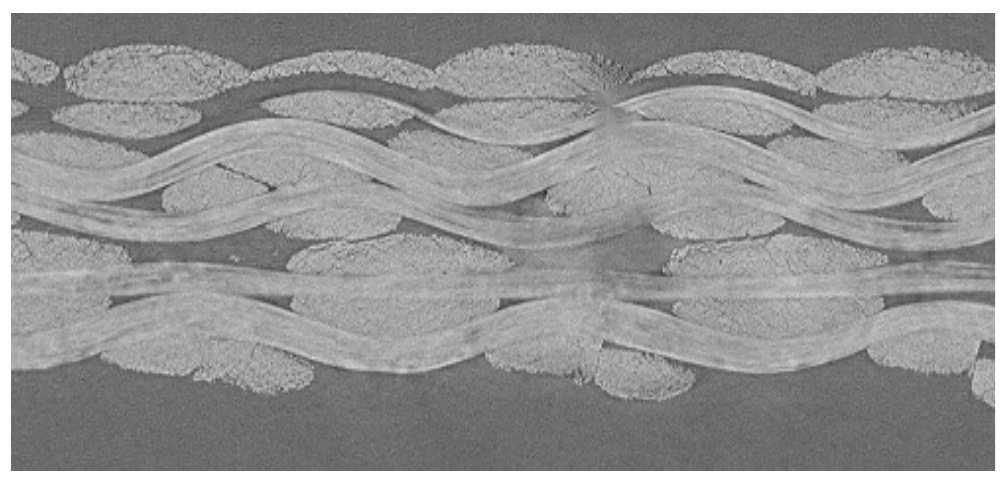}
     	\caption{Nominal frame.}
     	\label{nominal}
     \end{subfigure}
     \begin{subfigure}[b]{0.45\textwidth}
     	\centering
     	\includegraphics[scale = 0.6]{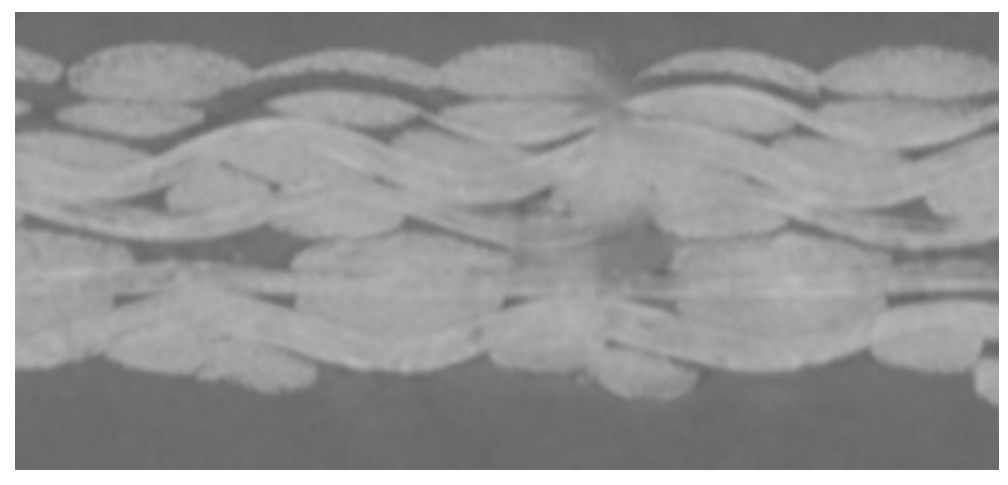}
     	\caption{Low contrast frame.}
     	\label{blurred}
     \end{subfigure}
     	\caption{A nominal contrast frame of the six ply weave (a) and a blurred, low contrast frame after a Gaussian kernel was applied (b). }
     \label{fig:comp}
\end{figure}

It is of additional interest how this proposed approach performs in a setting where texture and thresholding methods struggle or simply fail.  For this reason, a Gaussian kernel is applied to the standard frames of the six ply weave to generate a blurred set where pixel-to-pixel textures and contrast differences are diminished (Fig. \ref{fig:comp}). This method is commonly referred to as Gaussian blurring \cite{gedraite}. The nominal image (Figure \ref{nominal}) is convolved using the Gaussian kernal. The Gaussian blur formula for the pixel values of a blurred image is described by the convolution 
\begin{equation}
I_{blurred} = G(x,y) * I_{nominal} ,
\end{equation}
where the Gaussian filter $G(x,y)$ is passed over each pixel in the nominal image, $I_{nominal}$, to produce a blurred image, $I_{blurred}$. The filter values are based on the distance from the centered pixel where $x$ is the horizontal distance from a pixel and $y$ is the vertical distance from a pixel. The Gaussian filter in 2D is given by
\begin{equation}
G(x,y) = \frac{1}{2 \pi \sigma^2}e^{-\frac{x^2+y^2}{2 \sigma^2}} .
\end{equation} 
Often referred to as the radius of the Gaussian kernal, $\sigma$ determines the size of the Gaussian distribution. For this effort, the value of $\sigma$ was set to 1.5; this value of $\sigma$ resulted in images with little distinction between instance borders without fully compromising the image quality (Figure \ref{blurred}).  This second dataset is used to train a separate set of neural network parameters, and the performance is compared to the parameters trained with the standard dataset. The same manually instanced ground truth dataset is used for training with the blurred dataset.

\section{Results}
\label{Results}

Following training of the network with the ADEPT material system, multiple numerical studies are conducted to evaluate the effectiveness and demonstrate the utility of this proposed approach. First the performance of the network is evaluated on test datasets using the PQ metric. Two test datasets are considered: one with a similar weave geometry seen during traing (six ply) and one with a weave geometry not seen during training (four ply).  Additionally, this was repeated using the blurred datasets to compare and contrast the performance of the network when only lower contrast datasets are available.  Second, the first study is repeated with the corrective recognition algorithm to assess the impact of this algorithm on the PQ metric.  Third, the utility of this approach is demonstrated from the final voxel representation resulting from the automated process methodology. Specifically, the ability to capture yarn paths, contact regions, and cross-sectional distributions in the dataset is presented.


\subsection{Training Performance}

\begin{figure}[h]
	\centering
	\includegraphics[trim=0 0 0 20,clip,scale=0.7]{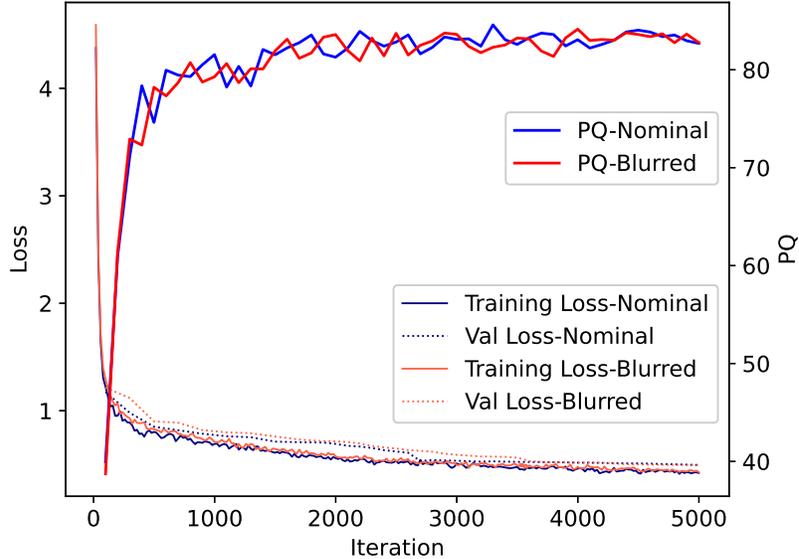}
	\caption{Training performance of the panoptic segmentation network for the nominal and blurred CT datasets. All evaluations are on the six ply dataset. PQ was evaluated on the validation set alone.}
	\label{Training}
\end{figure}

Both training loss and validation loss for each network were tracked to evaluate overfitting. The PQ of the validation sets were tracked to assess network perform at each iteration. Two panoptic segmentation networks were trained on the six ply dataset: one with Gaussian blurring (blurred) and one without (nominal). Training to 5000 iterations with a 4 image batch size took approximately 2 hours at the NASA Advanced Supercomputing facility with a single NVIDIA Volta V100 GPU. 

Overfitting does not occur during the training process (Fig. \ref{Training}). The validation loss is similar to the training loss for both networks thus revealing that the trained network generalizes well to the validation set. The panoptic segmentation network trained on the blurred dataset shows little to no significant difference in PQ score from the nominal dataset. Apart from the validation set, the PQ metric is evaulated in more depth on the test datasets in the following studies.

\subsection{PQ Performance without Corrective Recognition}
The PQ metric reveals how well the predictions from a trained panoptic segmentation network align with the ground truth. The PQ metrics for both trained networks were evaluated using both the six ply and four ply test datasets.  These metrics are additionally decomposed into the segmentation quality (SQ) and recognition quality (RQ) metrics (Table \ref{tab:PQ1}). Performance on the four ply dataset reveals each network’s ability to extrapolate to a similar composite textile material system. RQ is the bottleneck for the PQ performance in all four cases. The metrics reveal that while the trained deep learners generalize well to the same material (new six ply frames), they do not extrapolate well to a material system with different textures, contrast, and geometry (four ply frames). When attempting to extrapolate to four ply frames, the RQ is significantly lower.  For the network trained and evaluated on the nominal dataset, the average RQ experienced a 83.5\% decrease in performance. Similarly, the network trained and evaluated on the blurred dataset experienced a 85\% decrease in average RQ.  The average SQ also experienced a decrease in performance, but not to the extent of the average RQ.

\begin{table}[h]
	\centering
	\begin{tabular}{c c | c c c }
	\hline
	\hline
	\textbf{Training Dataset} & \textbf{Test Dataset}	& $PQ_{avg}$ &	$SQ_{avg}$ & $RQ_{avg}$ \\
	\hline
	Six ply – nominal & Six ply – nominal & 83.5 & 85.6 & 97.5 \\
	Six ply – nominal & Four ply – nominal & 10.8 & 62.7 & 16.1\\
	Six ply – Blurred & Six ply – Blurred & 81.8 & 84.1 & 97.2\\
	Six ply – Blurred & Four ply – Blurred & 9.92 & 67.9 & 14.6\\
	\hline
	\hline
	\end{tabular}
	\centering
	\caption{Performance metrics for both panoptic segmentation networks without the corrective methodology. The subscript $avg$ denotes the average across all frames in the dataset.}
	\label{tab:PQ1}
\end{table}

\subsection{PQ Performance with Corrective Recognition}

The proposed corrective recognition algorithm was applied to improve the RQ metric as noted in Table 2.  Because the corrective procedure purely addresses false positives and false negatives without significantly affecting segmented pixels, the RQ value is improved and the SQ remained the same. Consequently, the PQ metric is also improved.  The proposed corrective procedure provides marginal improvements ($<$1\%) for the panoptic segmentation network trained evaluated on the six ply dataset.  Because the initial RQ performance on the four ply dataset was poor, the average RQ experiences a more significant improvement (20-27\%) when the ply four prediction datasets are post-processed by the corrective methodology. However, even after correction, the average RQ remains poor.

\begin{table}[h]
	\centering
	\begin{tabular}{c c | c c c c }
	\hline
	\hline
	\textbf{Training Dataset} & \textbf{Test Dataset}	& $PQ_{avg,corr}$  & $RQ_{avg,corr}$\\
	\hline
	Six ply – nominal & Six ply – nominal & 84.1  & 98.2\\
	Six ply – nominal & Four ply – nominal & 13.1  & 20.8\\
	Six ply – Blurred & Six ply – Blurred & 81.9  & 97.4\\
	Six ply – Blurred & Four ply – Blurred & 12.6  & 18.6\\
	\hline
	\hline
	\end{tabular}
	\centering
	\caption{Performance metrics for both panoptic segmentation networks with the corrective methodology.}
	\label{tab:PQ2}
\end{table}

\subsection{Automated Processing Utility}  

\begin{figure}[h]
	\centering
	\includegraphics[width=0.5\textwidth]{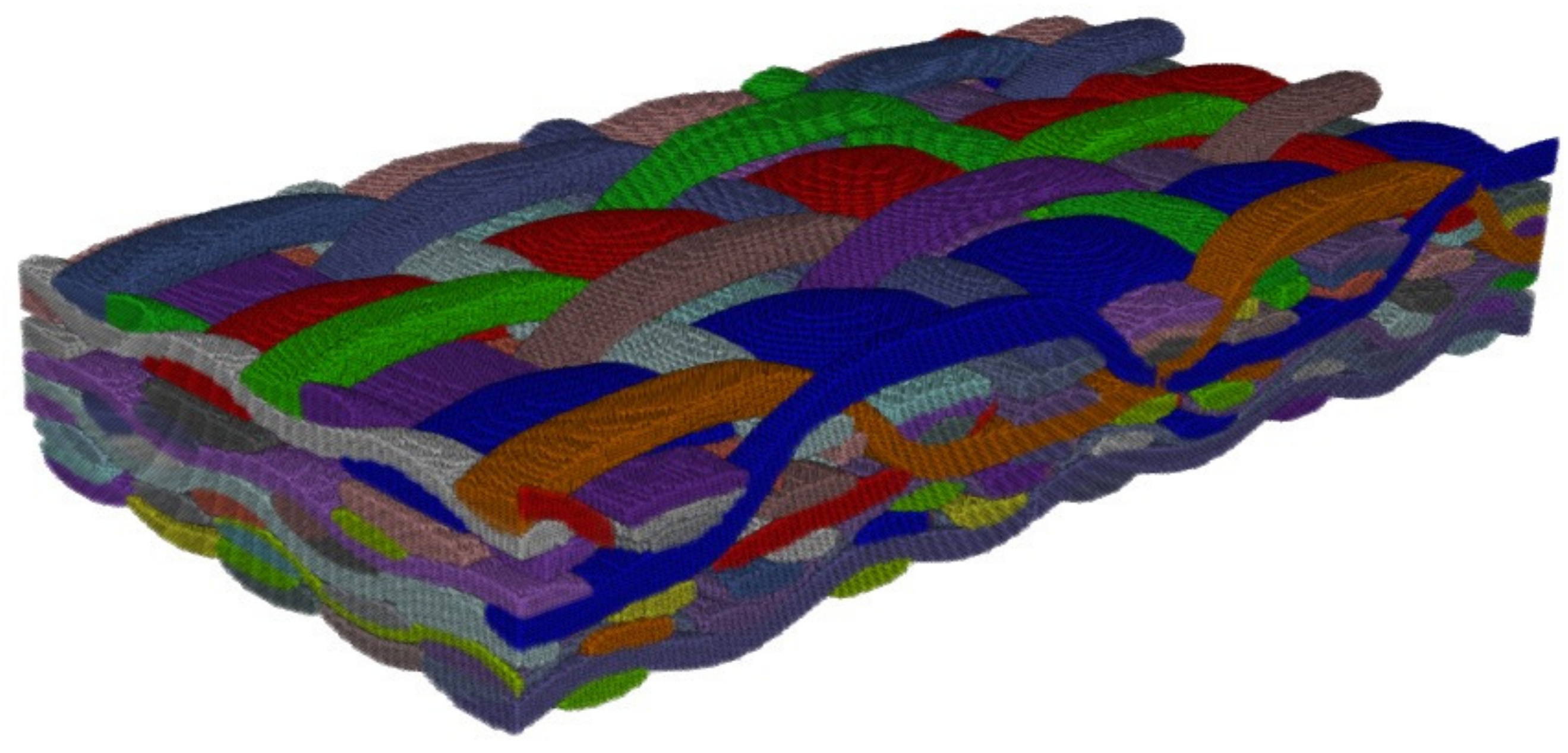}
	\caption{A six ply geometry segmented and reconstructed via panoptic segmentation predications and with post processing (both false positive removal and false negative corrections). Each unique yarn bundle is characterized by a unique color identifier.}
	\label{fig:segmented}
\end{figure}

To demonstrate the utility of this approach, a final voxel representation resulting from the automated process methodology is examined. After panoptic segmentation predictions were cast for individual frames, tracking was implemented to group like instances of yarns, the corrective procedure was applied, and a 3D geometry was assembled as shown in Fig. \ref{fig:segmented}. This representation corresponds to the PQ performance metric on row one of Table 2. Because recognition is consistent (the corrected RQ metric is 98.2), this automated processing approach is able to successfully identify each individual yarn in the CT data and maintain each yarns continuity in the depth direction.

\begin{figure}[h]
     \centering
     \begin{subfigure}[b]{0.48\textwidth}
     	\centering
     	\includegraphics[scale = 0.24]{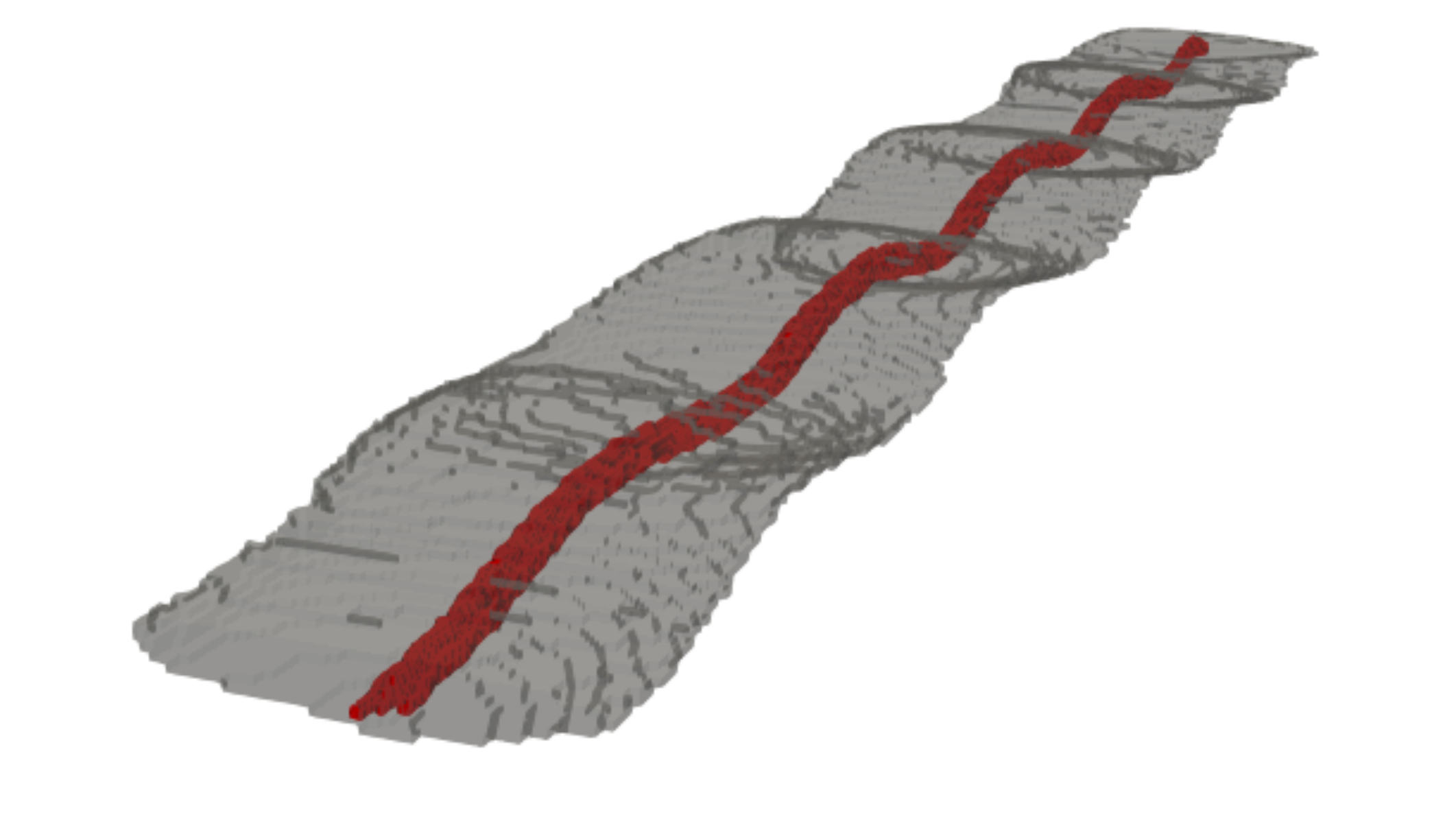}
     	\caption{Streamlines depict the yarn path of a segmented yarn.}
     	\label{vector}
     \end{subfigure}
     \begin{subfigure}[b]{0.48\textwidth}
     	\centering
     	\includegraphics[scale = 0.46]{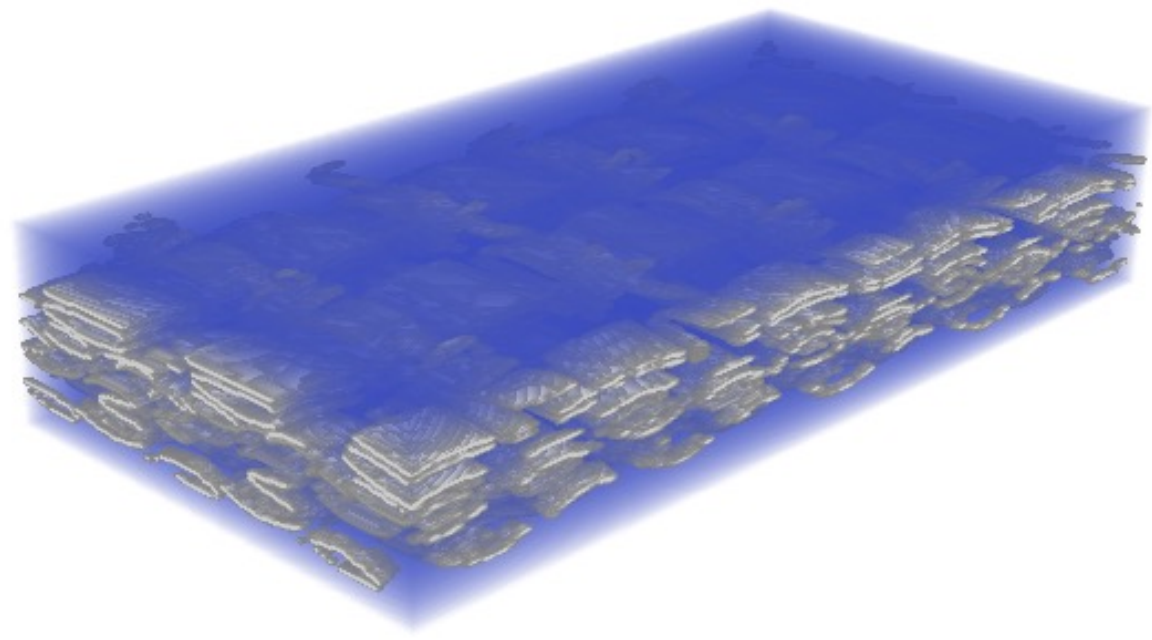}
     	\caption{The contact regions between yarns highlighted in gray}
     	\label{contact}
     \end{subfigure}
     	\caption{Two forms of additional model utility afforded by panoptic segmentation of the mesoscale.}
     \label{fig:instanced}
\end{figure}

Panoptic segmentation on the mesoscale affords the capability of modeling physics not possible with semantic segmentation.  Two modeling capabilities afforded over semantic segmentation are demonstrated here: determining individual yarn paths and contact regions between yarns.  Segmentation of each individual yarn permits tracking the yarns in space.  The yarn paths are denoted by the streamlines in Fig. \ref{vector}. Moreover, segmentation of each individual yarn allows for contact regions to be identified between individual yarns. Capturing these contact regions enable modeling of the friction between individual yarns as well as the matrix, if present. The contact regions are displayed in in Fig. \ref{contact}. Both anistropic material properties and friction contribute to the composite textile properties on the macroscale. 

\begin{figure}[h]
	\centering
	\includegraphics[trim=10 0 10 20,clip,width=0.5\textwidth]{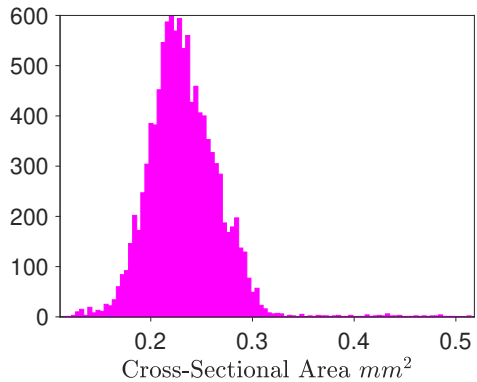}
	\caption{The cross-sectional areas of each individual yarn in each frame were collected to quantify the non-deterministic variations.}
	\label{fig:Distribution}
\end{figure}

Once the composite textile mesoscale is constructed via the proposed procedure, the reconstructed geometry can be probed for characterizing the geometric variations within.  A third utility afforded by this proposed approach over semantic segmentaion is the ability to characterize the yarn cross-sectional area variations in the dataset. The resulting histogram is displayed in Fig. \ref{fig:Distribution}.  

\section{Discussion}
\label{Disc}

The recent advent of panoptic segmentation networks affords new possibilities for modeling composite textiles from CT scans.  To the authors knowledge, this is the first automated processing approach capable of instance segmenting woven composite textiles.  The utility of this approach was captured in the following ways.  

	 Once trained on a material system, the panoptic segmentation network generalizes well to new frames of the same material system. One potential pitfall of this approach is that it does not extrapolate well to a dataset with a different geometry, contrast, and textures when trained on a single material system (in this case, just the six ply ADEPT material system).  This was demonstrated by evaluating performance on the four ply weave.  In future efforts, the transferability to new layups should be examined by training the deep learner on multiple woven composite textile layups. This transferability is desirable because training requires ground truth data, but the ability to generalize new frames of the same material system is still important as this ability precludes manually instancing the full dataset. Furthermore, it was demonstrated that panoptic segmentation can be utilized on a low-contrast dataset generated via Gaussian blurring. Prior thresholding and texture-sensitive segmentation approaches traditionally utilized to segment composite textiles struggle to segment low contrast datasets [15,16,18], so this is a promising advantage.
	
	Beyond merely assembling a geometry for modeling, it is desirable to know how representative, or consistent, the segmented geometry is to the desired geometry.  The panoptic segmentation-based approach presented here is a desirable improvement to this end. The PQ metric weighs both the ability to identify an instance (RQ) and the quality of the semantically segmented instance (SQ).  This metric is well suited for evaluating the reconstructed geometry when ground truth data is available for the evaluation.

	Identifying unique yarn bundles via instance segmentation enable higher fidelity modeling for computational analysis. Higher fidelity physical simulations corresponding to continuum mechanics and heat transfer can be conducted once the yarn paths and contract regions are identified. Common composite textiles used in aerospace applications are anisotropic at the mesoscale \cite{naouar1}. Identifying individual yarn paths permits assigning anistropic constitutive properties on this scale. In turn, this affects the resultant macroscopic material properties after homogenization. Contact phenomena, such as friction between individual yarns as well as the fiber reinforcement between a matrix, if present, contribute to the macroscopic constitutive properties. The surface area of the contact regions between yarns, as well as the matrix, if present, influence thermal phenomena. It is this additional capability that makes panoptic segmentation more desirable than semantic segmentation for computational simulations.  Furthermore, individual yarn identification lends to the non-deterministic characterization of the mesoscale geometry.  Prior to this work, the effort described in \cite{huang} conducted an in-depth characterization of the yarn geometries after manual segmentation. The approach presented here enables the characterization of yarn geometry profiles via panoptic segmentation instead of manual segmentation. These profiles are useful for quantifying the variations in the weaves, creating similar, synthetic weaves, and performing reduced order modeling \cite{bostanabad,han}.

\section{Conclusions}

A new approach for generating 3D digital representations of composite weaves geometries from CT images via deep learning is proposed.  This approach improves over existing methods by utilizing panoptic segmentation for the automated segmentation of woven composite textiles. Panoptic segmentation enables the identification of unique yarn instances within a frame, and it can be adopted for segmenting low-contrast datasets. This was demonstrated by evaluating two panoptic segmentation networks: one trained on nominal CT images and one trained on the same images processed with a Gaussian blur. 

After identifying unique yarn instances, an intersection-over-union based objected tracking method was adopted from video panoptic segmentation.  This tracking method enables continuity of unique yarn predictions through subsequent frames.  In turn, a 3D voxel representation of the mesoscale geometry was assembled.  From this assembled mesosclae geometry, yarn paths and contract regions were identified to demonstrate the utility of a panoptic segmentation based automated processing approach for higher fidelity physics models than what is possible via semantic segmentation.  Moreover, this approach is advantageous over semantic segmentation in that it permits the quantification of non-deterministic spatial variances of the yarns. This was demonstrated by quantifying the distribution of the yarn cross-sectional areas.  

To evaluate the reconstructed geometry from CT scans, the performance of the panoptic segmentation instance predictions were evaluated via the PQ metric. A corrective procedure to reduce false positives and false negatives was implemented to provide a boost to the RQ and consequently the PQ of the predictions.  While panoptic segmentation does not guarantee transferability to different weave mesostructures, it does generalize well to similar weave mesostructures with CT scans of similar contrast.  This evaluation was made by conducting predictions on CT frames not seen by the network during training.  The network generalized well to new frames of the same weave used for training (six ply) but performed poorly on the different layup weave with differing textures and contrast (four ply).

Overall, the panoptic segmentation based automated processing approach presented herein provides a new capabilities not afforded by existing segmentation and automated processing approaches. In future work, this approach should be extended to training datasets that include multiple woven composite textile layups and/or material systems, and the performance should be reevaluated. This future research is necessary to determine the extent that this approach can generalize to CT images of differing geometry, texture, and contrast.

\section{Acknowledgments}

This work was supported by a NASA Space Technology Graduate Research Opportunities Award, the Entry Systems Modeling Project under the NASA Game Changing Development Program, and the DARPA Transformative Design Program. Resources supporting this work were provided by the NASA High-End Computing (HEC) Program through the NASA Advanced Supercomputing (NAS) Division at Ames Research Center.

\section{Conflicts of Interest}
The authors declare that they have no conflict of interest. The algorithmic settings for all problems, as well the files with the final designs, are available upon request. 

\section{Supplementary Information}
The opinions and conclusions presented in this paper are those of the authors and do not necessarily reflect the views of the sponsoring organization.


\appendix;
\section{Network Backbone}
The ResNet-101-FPN network is utilized as the backbone for this effort \cite{he2}.  The layers utilized in ResNet-101 network are described in the table below. This network is comprised of repeating convolutional bottleneck blocks (BKB) resulting in 101 convolutional layers and a final, fully connected layer with a softmax classifier. The architecture is presented in Table \ref{tab:arc}. More detail on the bottlekneck blocks for deep residual networks as well as a diagrams for ResNet architectures can be found in \cite{he2}. The ResNet-101 architecture is then converted into a feature pyramid network (FPN) to provide the main backbone of the panoptic segmentation network. More on feature pyramid networks can be found in \cite{lin,kirillov2}. 
\label{AppA}

\begin{table}[h] 
	\centering
	\begin{tabular}{c | c c c c }
	\hline
	\hline
	\textbf{Layer Name} 	& Description & Repeat \\
	\hline
	conv1  & 7x7, 64, stride 2 & 1\\
	\hline
	conv2 & 3x3, max pool, stride 2 & 1\\
	\hline
	conv2 BKB  &  \shortstack{ 1x1, 64 \\ 3x3, 64 \\ 1x1, 256} & 3\\
	\hline
	conv3 BKB & \shortstack{ 1x1, 128 \\ 3x3, 128 \\ 1x1, 512} & 4\\
	\hline
	conv4 BKB & \shortstack{ 1x1, 256 \\ 3x3, 256 \\ 1x1, 1024} & 23\\
	\hline
	conv5 BKB & \shortstack{ 1x1, 512 \\ 3x3, 512 \\ 1x1, 2048} & 3\\
	\hline
output	 & fully connected and softmax  & 1\\
	\hline
	\hline
	\end{tabular}
	\centering
	\caption{Res-Net-101 network architecture. The 'description' denotes the filter size, number of filters, and stride if not one. The first layer of conv3 BKB, conv4 BKB and conv 5 BKB are downsampled with a stride of 2.}
	\label{tab:arc}
\end{table}

\bibliographystyle{elsarticle-num} 
\bibliography{PS}


%
%
\end{document}